\documentclass[10pt,twocolumn,letterpaper]{article}
\usepackage{multirow}
\usepackage{booktabs}
\usepackage{cvpr}
\usepackage{times}
\usepackage{epsfig}
\usepackage{graphicx}
\usepackage{amsmath}
\usepackage{amssymb}
\usepackage{algorithm}
\usepackage{algorithmic}
\usepackage{balance}

\usepackage[breaklinks=true,bookmarks=false]{hyperref}

\cvprfinalcopy 

\def\cvprPaperID{2722} 

\begin{document}

\title{Defense against Universal Adversarial Perturbations}

\author{Naveed Akhtar* \hspace{4mm} Jian Liu*  \hspace{4mm} Ajmal Mian\\
{\small*The authors contributed equally to this work.}\\
School of Computer Science and Software Engineering\\
The University of Western Australia\\
{\tt\small naveed.akhtar@uwa.edu.au,}
{\tt\small jian.liu@research.uwa.edu.au,}
{\tt\small ajmal.mian@uwa.edu.au}
}

\maketitle

\begin{abstract}
Recent advances in Deep Learning show the existence of image-agnostic quasi-imperceptible perturbations that when applied to `any' image  can fool a state-of-the-art network classifier to change its prediction about the  image label. These `Universal Adversarial Perturbations' pose a serious threat to the success of Deep Learning in practice.  We present the first dedicated framework to effectively defend the networks against such perturbations. Our approach learns a Perturbation Rectifying Network (PRN) as `pre-input' layers to a targeted model, such that the targeted model needs no modification. The PRN is learned from real and synthetic image-agnostic perturbations, where an efficient method to compute the latter is also proposed. A perturbation detector is separately trained on the Discrete Cosine Transform of the input-output difference of the PRN.   A query image is first passed through the PRN and verified by the detector. If a perturbation is detected, the output of the PRN is used for label prediction instead of the actual image. A rigorous evaluation shows that our framework can defend the  network classifiers against  unseen adversarial perturbations in the real-world scenarios  with up to $97.5\%$ success rate. The PRN also generalizes well in the sense that training for one targeted  network defends another network with a comparable success rate.

\end{abstract}
\vspace{-4mm}
\section{Introduction}
\label{sec:Intro}
\begin{figure}[t] 
   \centering
   \includegraphics[width=2.7in]{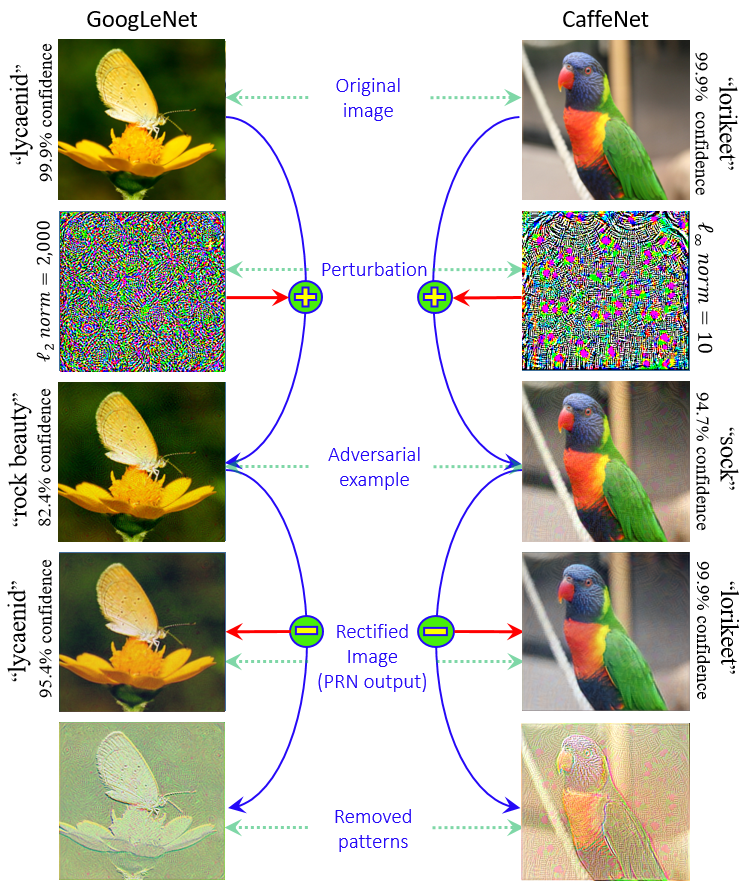} 
   \caption{Adding quasi-imperceptible universal adversarial perturbations~\cite{UniAdPert2017CVPR}  can fool neural networks. The proposed framework rectifies the images to restore the network predictions. The patterns  removed by rectification are separately analyzed to decide on the presence of adversarial perturbations in images. The shown  `perturbations' and `removed patterns' are normalized on different scales for better visualization.}
   \label{fig:Teaser}
   \vspace{-4mm}
\end{figure}

Deep Neural Networks are at the heart of the current advancements in Computer Vision and Pattern Recognition, providing state-of-the-art performance on many challenging classification tasks~\cite{goodfellow2016deep}, \cite{he2016deep}, \cite{huang2016densely},  \cite{krizhevsky2012imagenet},  \cite{simonyan2014very}, \cite{szegedy2015going}.
However, Moosavi-Dezfooli et al.~\cite{UniAdPert2017CVPR} recently showed the possibility of fooling the deep networks to change their prediction about `any' image that is slightly perturbed with the  \emph{Universal Adversarial Perturbations}.
For a given network model, these image-agnostic (hence~\emph{universal}) perturbations can be computed  rather easily~\cite{UniAdPert2017CVPR}, \cite{moosavi2017analysis}. 
The perturbations remain quasi-imperceptible (see Fig.~\ref{fig:Teaser}), yet the \emph{adversarial examples} generated by adding the perturbations to the images fool the networks with alarmingly high probabilities~\cite{UniAdPert2017CVPR}. 
Furthermore, the fooling is able to generalize well across different network models. 

Being image-agnostic, universal adversarial perturbations can be conveniently exploited to fool models on-the-fly on unseen images by using pre-computed perturbations. This even eradicates the need of on-board computational capacity that is needed for generating image-specific perturbations~\cite{fawzi2016robustness}, \cite{lu2017no}.
This fact, along the cross-model generalization of universal perturbations make  them particularly relevant to the practical cases where a model is deployed in a possibly hostile environment.
Thus, defense against these perturbations is a necessity for the success of Deep Learning in practice.
The need for  counter-measures against these perturbations becomes even more pronounced considering  that the real-world scenes (e.g.~sign boards on roads) modified by the adversarial perturbations can also behave as adversarial examples for the networks~\cite{kurakin2016adversarial}.

This work proposes the first dedicated defense against  the universal adversarial perturbations~\cite{UniAdPert2017CVPR}. The major contributions of this paper are as follows:
\begin{itemize}
\item We propose to learn a Perturbation Rectifying Network (PRN) that is trained as the `pre-input' of a targeted network model. This allows our framework to  provide defense to already deployed networks without the need of modifying them.
\item We propose a method to efficiently compute synthetic image-agnostic adversarial perturbations to effectively train the PRN. The successful generation of these perturbations complements the theoretical findings of  Moosavi-Dezfooli~\cite{moosavi2017analysis}.
\item We also propose a separate perturbation detector that is learned from the Discrete Cosine Transform of the image rectifications performed by the  PRN for clean and perturbed examples. 
\item Rigorous evaluation is performed by defending the GoogLeNet~\cite{szegedy2015going}, CaffeNet~\cite{krizhevsky2012imagenet} and VGG-F network~\cite{chatfield2014return}\footnote{The choice of the networks is based on the computational feasibility of generating the adversarial perturbations for the evaluation protocol in Section~\ref{sec:Exp}. However, our approach is generic in nature.}, demonstrating up to $97.5\%$ success rate on unseen  images possibly modified with unseen perturbations. Our experiments also show that the proposed PRN generalizes well across different network models.
\end{itemize}

\section{Related work}
\label{sec:RW}
The robustness of image classifiers against adversarial perturbations has gained significant attention in the last few years~\cite{fawzi2015analysis}, \cite{fawzi2016robustness}, \cite{nguyen2015deep}, \cite{rozsa2016adversarial},  \cite{sabour2015adversarial}, \cite{sharif2016accessorize}, \cite{tabacof2016exploring}. 
Deep neural networks became the center of attention in this area after  
Szegedy et al.~\cite{szegedy2013intriguing} first demonstrated the existence of adversarial perturbations for such networks. 
See \cite{survey} for a recent review of literature in this direction.
Szegedy et al.~\cite{szegedy2013intriguing} computed adversarial  examples for the networks by adding quasi-imperceptible perturbations to the images, where the perturbations were estimated by maximizing the network's prediction error.
Although these perturbations were image-specific, it was shown that the same perturbed images were able to fool multiple network models.
Szegedy et al.~reported encouraging results for improving the model robustness against the adversarial attacks by using adversarial examples for training, a.k.a.~\emph{adversarial training}.

Goodfellow et al.~\cite{goodfellow2014explaining} built on the findings in~\cite{szegedy2013intriguing} and developed a `fast gradient sign method' to efficiently generate adversarial examples that can be used for  training the networks.
They hypothesized that it is the linearity 
of the deep networks that makes them vulnerable to the adversarial perturbations. 
However, Tanay and Griffin~\cite{tanay2016boundary} later constructed the image classes that do not suffer from the adversarial examples for the linear classifiers. Their arguments about the existence of the adversarial perturbations again point  towards the over-fitting phenomena, that can be alleviated by regularization.
Nevertheless, it remains unclear how a network should be regularized for robustness against adversarial examples. 

Moosavi-Dezfooli~\cite{moosavi2016deepfool} proposed the DeepFool algorithm to compute image-specific adversarial perturbations by assuming that the loss function of the network is linearizable around the current training sample.
In contrast to the one-step perturbation estimation~\cite{goodfellow2014explaining}, their approach  computes the perturbation in an iterative manner.  
They also reported that augmenting training data with  adversarial examples significantly increases the robustness of networks against the  adversarial perturbations.
Baluja and Fischer~\cite{baluja2017adversarial} trained an Adversarial Transformation Network to generate adversarial examples against a target network.
Liu et al.~\cite{liu2016delving} analyzed the transferability of adversarial examples. They studied this property for both targeted and non-targeted examples, and proposed an ensemble based approach to generate the  examples with better transferability.

The above-mentioned techniques mainly focus on  generating adversarial examples, and address the defense against those examples with adversarial training. 
In-line with our take on the problem, few recent techniques also directly focus on the defense against the adversarial examples. 
For instance, Lu et al.~\cite{luo2015foveation} mitigate the issues resulting from the adversarial perturbations using foveation. Their main argument is that the neural networks (for ImageNet~\cite{russakovsky2015imagenet}) are robust to the foveation-induced scale and translation variations of the images, however, this property does not generalize to the perturbation transformations.

Papernot et al.~\cite{papernot2016distillation} used distillation~\cite{hinton2015distilling} to make the neural networks more robust against the adversarial perturbations. However, Carlini and Wagner~\cite{carlini2017towards} later introduced adversarial attacks that can not be defended by the distillation method.  
Kurakin et al.~\cite{kurakin2016adversarial1} specifically studied the adversarial training for making large models (e.g.~Inception v3~\cite{szegedy2016rethinking}) robust to perturbations, and found that the training indeed provides robustness against the perturbations generated by the one-step methods~\cite{goodfellow2014explaining}. However, Tramer et al.~\cite{tramer2017ensemble} found that  this robustness weakens for the adversarial examples learned using different networks i.e.~for the black-box attacks~\cite{liu2016delving}. Hence, ensemble adversarial training was proposed in~\cite{tramer2017ensemble} that uses adversarial examples generated by multiple networks.

Dziugaite et al.~\cite{dziugaite2016study} studied the effects of JPG compression on  adversarial examples and found that the compression can sometimes revert network fooling.
Nevertheless, it was concluded that JPG compression alone is insufficient as a defense against adversarial attacks. Prakash et al.~\cite{prakash2018deflecting} took advantage of localization of the perturbed pixels in their defense.
Lu et al.~\cite{lu2017safetynet} proposed SafetyNet for detecting and rejecting adversarial examples for the conventional network classifiers (e.g.~VGG19~\cite{he2015delving})  that capitalizes on the late stage ReLUs of the network to detect the perturbed examples.
Similarly, a proposal of appending the deep neural networks with detector subnetworks was also presented by Metzen et al.~\cite{metzen2017detecting}. 
In addition to the classification, adversarial examples and robustness of the deep networks against them have also been recently investigated for the tasks of semantic segmentation and object detection~\cite{fischer2017adversarial}, \cite{lu2017no}, \cite{xie2017adversarial}. 

Whereas the central topic of all the above-mentioned literature is the perturbations computed for \emph{individual} images, Moosavi-Dezfooli~\cite{UniAdPert2017CVPR} were the 
 first to show the existence of image-agnostic perturbations for neural networks. These  perturbations were further analyzed in \cite{moosavi2017analysis}, whereas Metzen et al.~\cite{metzen2017universal} also showed their existence for semantic image segmentation.
To date, no dedicated technique exists for defending the networks against the universal adversarial perturbations, which is the topic of this paper.

\section{Problem formulation}
\label{sec:PF}
Below, we  present the notions of \emph{universal adversarial perturbations} and the  \emph{defense} against them more formally. 
Let $\boldsymbol\Im_c \in \mathbb R^{\text d}$ denote the distribution of the (clean) natural images in a d-dimensional space, such that, a class label is associated  with its every sample ${\bf I}_c \sim \boldsymbol\Im_c$. 
Let $\mathcal C(.)$ be a classifier (a deep network) that maps an image to its class label, i.e.~$\mathcal C({\bf I}_c): {\bf I}_c \to \ell \in \mathbb R$.
The vector  $\boldsymbol \rho \in \mathbb R^{\text d}$ is a universal adversarial perturbation for the classifier, if it satisfies the following constraint:
\begin{align}
\underset{{\bf I}_c \sim \boldsymbol\Im_c}{\mathrm{\text P}} \Big( \mathcal C({\bf I}_c) \neq \mathcal C({\bf I}_c + \boldsymbol \rho) \Big) \geq \delta ~~~\text{s.t.}~~||\boldsymbol\rho||_p \leq \xi,
\label{eq:def1}
\end{align}
where P(.) is the probability, $||.||_p$ denotes the $\ell_p$-norm of a vector such that $p \in [1, \infty)$, $\delta \in (0, 1]$ denotes the \emph{fooling ratio}  and $\xi$ is a pre-defined constant. 
In the text to follow, we alternatively refer to $\boldsymbol\rho$ as  the \emph{perturbation} for brevity.

In~(\ref{eq:def1}), the perturbations in question are \emph{image-agnostic}, hence Moosavi-Dezfooli et al.~\cite{UniAdPert2017CVPR} termed them \emph{universal}\footnote{A single perturbation that satisfies (\ref{eq:def1}) for \emph{any} classifier is referred as `doubly universal' by Moosavi-Dezfooli et al.~\cite{UniAdPert2017CVPR}. We focus on the singly universal perturbations in this work.}. 
According to the stated definition, the parameter $\xi$ controls the norm of the perturbation. 
For the \emph{quasi-imperceptible} perturbations, the value of this parameter should be very small as compared to the image norm $||{\bf I}_c||_p$. On the other hand, a larger $\delta$ is  required for the perturbation to fool the classifier with a higher  probability. In this work, we let $\delta \geq 0.8$ and consider the perturbations constrained by their $\ell_2$ and $\ell_{\infty}$ norms.
For the $\ell_2$-norm, we let $\xi = 2,000$, and select $\xi = 10$ for the $\ell_{\infty}$-norm perturbations.
For both types, these values are $\sim 4\%$ of the means of the respective image norms  used in our experiments (in Section~\ref{sec:Exp}), which is the same as~\cite{UniAdPert2017CVPR}. 

To defend $\mathcal C(.)$ against the  perturbations, we seek two components of the  defense mechanism. (1) A perturbation `detector' $\mathcal D({\bf I}_{\boldsymbol\rho/ c}): {\bf I}_{\boldsymbol\rho/ c} \rightarrow [0,1]$ and (2) a perturbation `rectifier' $\mathcal R({\bf I}_{\boldsymbol\rho}): {\bf I}_{\boldsymbol\rho} \rightarrow \widehat{\bf I}$, where ${\bf I}_{\boldsymbol\rho} = {\bf I}_c + \boldsymbol\rho$.
The detector determines whether an unseen image ${\bf I}_{\boldsymbol\rho/ c}$ is perturbed or clean.
The objective of the rectifier is to compute a transformation $\widehat{{\bf I}}$ of the perturbed image such that $\underset{{\bf I}_c \sim \boldsymbol\Im_c}{\mathrm{\text P}} \Big( \mathcal C({\widehat{\bf I}}) = \mathcal C({\bf I}_c) \Big) \approx 1$. 
Notice that the rectifier does not seek to improve the prediction of $\mathcal C(.)$ on the rectified version of the image beyond the classifier's performance on the clean/original image.  
This ensures  stable induction of $\mathcal R(.)$. Moreover, the formulation allows us to compute $\widehat{{\bf I}}$ such that $||\widehat{\bf I} - {\bf I}_c ||_2 > 0$.
We leverage this property to learn $\mathcal R(.)$ as the pre-input layers of $\mathcal C(.)$ in an end-to-end fashion.


\begin{figure*}[t] 
   \centering
   \includegraphics[width=5.3in]{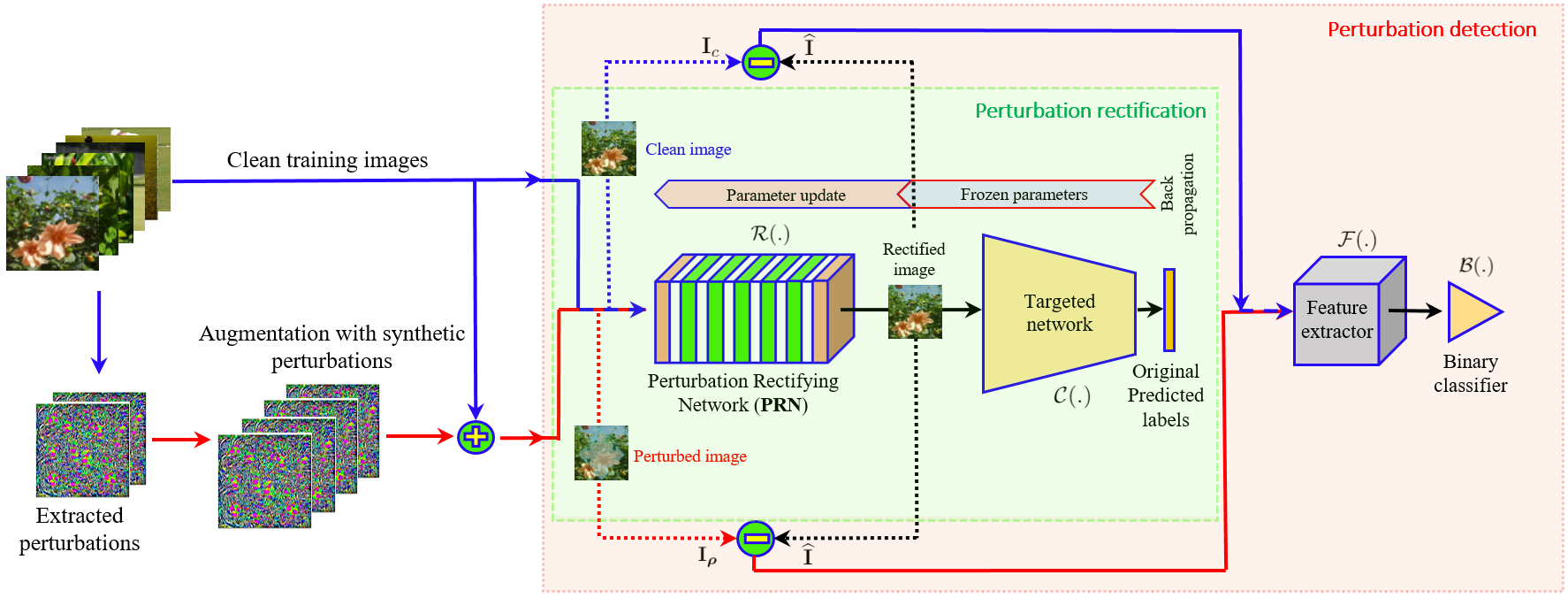} 
   \caption{Training schematics: From the clean data, image-agnostic perturbations are computed and augmented with the synthetic perturbations. Both clean and perturbed  images are fed to the Perturbation Rectifying Network (PRN). The PRN is learned by attaching it to the first layer of the targeted network such that the parameters of the targeted network are kept frozen during the PRN training. The perturbation detection mechanism extracts discriminative features from the difference between the inputs and outputs of the PRN and learns a binary classifier. To classify an unseen test image ${\bf I}_{\boldsymbol\rho/c}$, first $\mathcal D({\bf I}_{\boldsymbol\rho/c}) = \mathcal B(\mathcal F( {\bf I}_{\boldsymbol\rho/c} -  \mathcal R({\bf I}_{\boldsymbol\rho/c})))$ is computed. If a perturbation is detected then $\mathcal R({\bf I}_{\boldsymbol\rho/c}) $ is used as the input to the classifier $\mathcal C(.)$ instead of the actual test image.}  
   \label{fig:Schema}
   \vspace{-3mm}
\end{figure*}

\section{Proposed approach}
\label{sec:PR}
We draw on the insights from the literature reviewed in Section~\ref{sec:RW} to develop a framework for defending a (possibly) targeted  network  model against universal adversarial perturbations.
Figure \ref{fig:Schema} shows the schematics of our approach  to learn the  `rectifier'  and the `detector'  components of the defense framework.
We use the  Perturbation Rectifying Network (PRN) as the `rectifier', whereas  a binary classifier is eventually trained to detect the adversarial perturbations in the  images.
The framework uses  both real and synthetic perturbations for  training. 
The  constituents  of the proposed  framework are explained below.

\subsection{Perturbation Rectifying Network (PRN)}
\label{sec:PRN}
At the core of our technique is the \emph{Perturbation Rectifying Network} (PRN), that is trained as pre-input layers to the targeted network classifier.  The PRN is attached to the first layer of the classification network and the joint network is trained to minimize the following cost: 
\begin{align}
\mathcal J(\boldsymbol\theta_p, {\bf b}_p) = \frac{1}{N} \sum\limits_{i=1}^{N} \mathcal{L}(\ell_i^*, {\ell}_i),
\label{eq:cost}
\end{align}
where $\ell^*_i$ and $\ell_i$ are the labels predicted by the joint network and the targeted network respectively,  such that $\ell_i$ is necessarily computed for the clean image.
For the $N$ training examples, $\mathcal L(.) $ computes the loss, whereas $\boldsymbol\theta_p$ and ${\bf b}_p$ denote the PRN weight and bias parameters.

In Eq.~(\ref{eq:cost}) we define  the cost  over the parameters of  PRN only, which  ensures that the (already deployed) targeted network does not require any modification for the defense being provided by our framework. 
This strategy is orthogonal to  the existing  defense techniques that  either update the targeted model using adversarial training to make the networks  more  robust~\cite{kurakin2016adversarial1}, \cite{tramer2017ensemble}; or incorporate architectural changes to the targeted network, which may include adding a subnetwork to the model~\cite{metzen2017detecting} or tapping into the  activations of certain layers to detect the adversarial examples~\cite{lu2017safetynet}.
Our defense mechanism acts as an external wrapper for the targeted network such that the PRN (and the detector) trained to counter the adversarial attacks can be kept secretive in order refrain from potential counter-counter attacks\footnote{PRN+targeted network are end-to-end differentiable and the joint network can be  susceptible to stronger attacks if PRN is not secretive. However, stronger perturbations are also more easily detectable by our  detector.}.  This is a highly desirable property of defense frameworks in the  real-world scenarios. 
Moosavi-Dezfooli~\cite{UniAdPert2017CVPR} noted that the universal adversarial perturbations can still exist for a model even after their  adversarial training. 
The proposed framework constitutionally caters for this problem.

We train the PRN using both clean and adversarial examples to ensure that the image transformation learned by  our network is not biased towards the  adversarial examples.
For training,   $\ell_i$ is computed separately with the targeted network for the clean version of the $i^\text{th}$ training example. 
The PRN is implemented as 5-ResNet blocks~\cite{he2016deep} sandwiched by convolution layers. 
The $224\times224\times 3$ input image is fed to Conv $3\times3$, stride = 1, feature maps = 64, `same' convolution; followed by 5 ResNet blocks, where each block consists of two convolution layers with  ReLU activations~\cite{nair2010rectified}, resulting in $64$ feature maps. The feature maps of the last ResNet block are processed by Conv  $3\times3$, stride = 1, feature maps = 16, `same' convolution; and then  Conv $3\times3$, stride = 1, feature maps = 3, `same' convolution. 

We use the cross-entropy loss~\cite{goodfellow2016deep} for training the PRN with the help of ADAM optimizer~\cite{kingma2014adam}.
The exponential decay rates for the first and the second moment estimates are set to 0.9 and 0.999 respectively. 
We set the initial learning rate to 0.01, and decay it by $10\%$ after each 1K iterations. 
We used mini-batch size of 64, and trained the PRN for a given targeted network for at least 5 epochs.

\subsection{Training data }
\label{sec:synt}
The PRN is trained using clean images as well as their adversarial counterparts, constructed by adding perturbations to the clean images.
We compute the latter by first generating a set of perturbations  $\boldsymbol\rho \in \mathcal P \subseteq \mathbb R^{\text d}$ following Moosavi-Dezfooli et al.~\cite{UniAdPert2017CVPR}. Their algorithm computes a universal perturbation in an iterative manner. 
In its inner loop (ran over the training images), the algorithm seeks a minimal norm vector~\cite{moosavi2016deepfool} to fool the network on a given image. 
The current estimate of $\boldsymbol\rho$ is updated by adding to it the sought vector  and back-projecting the resultant vector onto the $\ell_p$ ball of radius $\xi$.
The outer loop ensures that the desired fooling ratio is achieved over the complete training set.
Generally, the algorithm requires several passes on the training data  to achieve an acceptable fooling ratio. We refer to~\cite{UniAdPert2017CVPR} for further details on the algorithm. 

A PRN trained with more adversarial patterns underlying the training images is expected to perform better.  
However, it becomes computationally infeasible to generate a large (e.g.~$>100$) number of perturbations using the above-mentioned algorithm. 
Therefore, we devise a mechanism to efficiently generate synthetic perturbations $\boldsymbol\rho_s \in \mathcal P_s \subseteq \mathbb R^{\text d}$ to augment the set of available perturbations for training the PRN.
The synthetic perturbations are computed using the set $\mathcal P$ while capitalizing on the theoretical results of \cite{moosavi2017analysis}.
To generate the synthetic perturbations, we compute the vectors that satisfy the following conditions: (c1) $\boldsymbol\rho_s \in \boldsymbol\Psi_{\mathcal P}^+: \boldsymbol\Psi_{\mathcal P}^+=$ positive orthant of the subspace spanned by the elements of ${\mathcal P}$. (c2) $|| \boldsymbol\rho_s ||_2 \approx  \mathbb E\left[ ||\boldsymbol\rho ||_2, \forall \boldsymbol\rho \in \mathcal P \right]$ and (c3)\footnote{For the perturbations restricted by their $\ell_2$-norm only, this condition is ignored. In that case, (c2) automatically  ensures $|| \boldsymbol\rho_s||_2 \approx \xi$.} $|| \boldsymbol\rho_s ||_{\infty} \approx \xi$.
The procedure for computing the synthetic perturbations that are constrained by their $\ell_{\infty}$-norm is summarized in Algorithm~\ref{alg:l_infty}. We refer to the supplementary material of the paper for the  algorithm to compute the $\ell_2$-norm perturbations. 

\newcommand{\isEq}[1]{\overset{#1}{\sim}}
\begin{algorithm}[t]
 \caption{$\ell_{\infty}$-norm synthetic perturbation generation}
 \label{alg:l_infty}
 \begin{algorithmic}[1]
 \renewcommand{\algorithmicrequire}{\textbf{Input:}}
 \renewcommand{\algorithmicensure}{\textbf{Output:}}
 \REQUIRE Pre-generated perturbation samples $\mathcal P \subseteq \mathbb R^{\text{d}}$, number of new samples to be generated $\eta$, threshold $\xi$.
 \ENSURE Synthetic perturbations $\mathcal{P}_s \subseteq \mathbb R^{d}$
 \STATE set $\mathcal{P}_s =\{\}$; $\ell_2$-threshold $= \mathbb E\left[ \{ ||\boldsymbol\rho_{i \in \mathcal P}||_2 \}_{i =1}^{|\mathcal P|} \right]$; \\$\mathcal P_n = \mathcal P$ with $\ell_2$-normalized elements.
 \WHILE {$|\mathcal{P}_s| < \eta$}
\STATE{set $\boldsymbol\rho_s = {\bf 0}$} 
 	\WHILE{$||\boldsymbol\rho_s||_{\infty} < {\xi}$} 
    	\STATE  $z\sim \text{unif}(0,1) \odot {\xi}  $
        \STATE $\boldsymbol\rho_s = \boldsymbol\rho_s + (z~~\odot \isEq{\text{rand}} \mathcal P_n) $
 	\ENDWHILE
    \IF {$||\boldsymbol\rho_s||_2 \geq \ell_2$-threshold}
    	\STATE $\mathcal P_s = \mathcal P_s \bigcup \boldsymbol\rho_s$
    \ENDIF
\ENDWHILE
  \STATE return 
 \end{algorithmic}
 \end{algorithm}

To generate a synthetic perturbation, Algorithm~\ref{alg:l_infty} searches for $\boldsymbol\rho_s$ in $\boldsymbol\Psi_{\mathcal P}^+$ by taking  small random steps in the directions governed by the unit vectors of the elements of $\mathcal P$.
The random walk continues until the $\ell_{\infty}$-norm of $\boldsymbol\rho_s$ remains smaller than $\xi$. 
The algorithm selects the found $\boldsymbol\rho_s$ as a valid perturbation if the $\ell_2$-norm of the vector is comparable to the Expected value of the $\ell_2$-norms of the vectors in $\mathcal P$.
For generating the $\ell_2$-norm perturbations, the corresponding algorithm given in the supplementary material  terminates the random walk based on $|| \boldsymbol\rho_s||_2$ in line-4, and directly selects the computed $\boldsymbol\rho_s$ as the desired perturbation.  
Analyzing the robustness of the deep networks against the universal  adversarial perturbations, Moosavi-Dezfooli~\cite{moosavi2017analysis} showed the existence of shared directions (across different data points) along which a decision boundary induced by a network becomes highly positively curved. 
Along these vulnerable directions,  small universal  perturbations  exist that can fool the network to change its predictions about the labels of the data points.
Our algorithms search for the synthetic perturbations along those directions, whereas the  knowledge of the desired directions is borrowed from $\mathcal P$.

\begin{figure}[t] 
   \centering
   \includegraphics[width=2in]{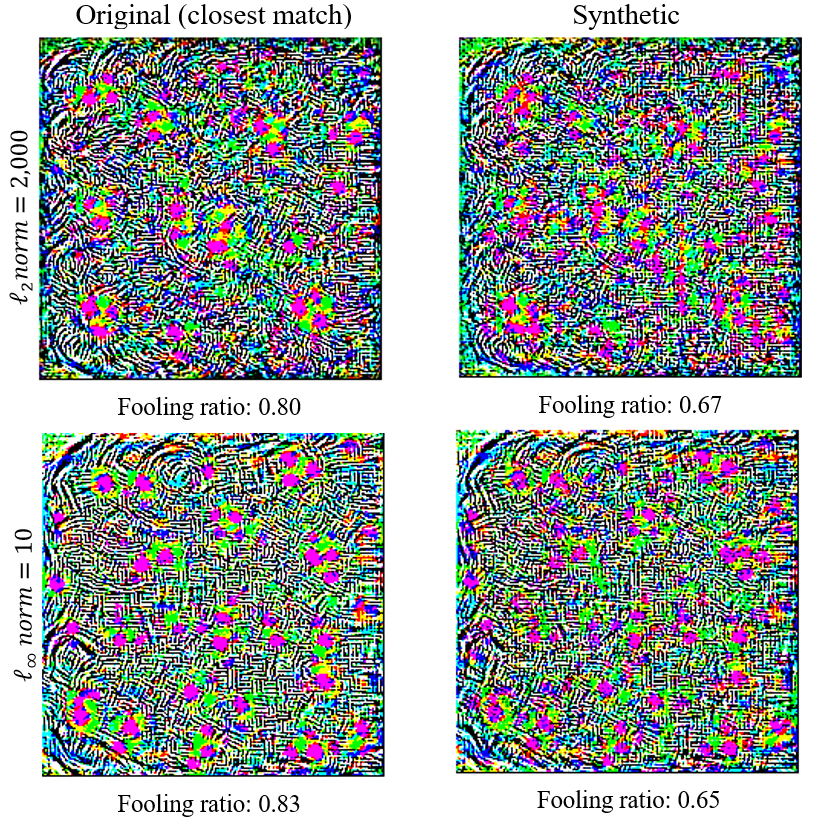} 
   \caption{Illustration of synthetic perturbations computed for the CaffeNet~\cite{krizhevsky2012imagenet}: The corresponding closest matches in set $\mathcal P$ are also shown. The dot product between the vectorized perturbations with their closest matches are 0.71 and 0.83 respectively for the $\ell_2$ and $\ell_{\infty}$-norm  perturbations.}
   \label{fig:Synth}
   \vspace{-3mm}
\end{figure}

Fig.~\ref{fig:Synth} exemplifies the typical synthetic perturbations generated by our algorithms for the $\ell_2$ and $\ell_{\infty}$ norms. It also shows the corresponding closest matches in the set $\mathcal P$ for the given perturbations. The fooling ratios for the synthetic perturbations is generally not as high as the original ones, nevertheless the values remain in an acceptable range.
In our experiments (Section~\ref{sec:Exp}), augmenting the training data with the synthetic perturbations consistently helped in early convergence and better performance of the PRN.  
We note that the acceptable fooling ratios demonstrated by the synthetic perturbations in this work complement the theoretical findings in~\cite{moosavi2017analysis}.   
Once the set of synthetic perturbations $\mathcal{P}_s$ is computed, we construct $\mathcal P^* = \mathcal P \bigcup \mathcal P_s$ and use it to perturb the  images in our training data.

\subsection{Perturbation detection}
\label{sec:PD}
While studying the JPG compression as a mechanism to mitigate the effects of the (image-specific) adversarial perturbations,  Dziugaite et al.~\cite{dziugaite2016study} also suggested the Discrete Cosine Transform (DCT) as a possible candidate to reduce the effectiveness of the  perturbations. Our experiments, reported in supplementary material,  show that the DCT based compression can also be exploited to reduce the network fooling ratios under the universal adversarial perturbations.
However, it becomes difficult to decide on the required compression rate, especially when it is not known whether the image in question is actually perturbed or not. Unnecessary rectification often leads to degraded performance of the networks on the clean images.  

Instead of using the DCT to remove the perturbations, we exploit it for perturbation detection in our approach. Using  the training data that contains both clean and perturbed images, say~${\bf I}_{\boldsymbol\rho/ c}^{\text{train}}$,  we first compute $\mathcal F( {\bf I}_{\boldsymbol\rho/ c}^{\text{train}} - \mathcal R({\bf I}_{\boldsymbol\rho/ c}^{\text{train}}))$ and then learn a binary classifier $\mathcal B(\mathcal F) \rightarrow [0, 1]$ with the data labels denoting the input being `clean' or `perturbed'.  We implement $\mathcal F(.)$ to compute the log-absolute values of the 2D-DCT coefficients of the gray-scaled image in the argument, whereas an SVM is learned as $\mathcal B(.)$. The function $\mathcal D(.) = \mathcal B(\mathcal F(.))$ forms the detector component of our defense framework. 
To classify a test image ${\bf I}_{\boldsymbol\rho/ c}$, we  first evaluate $\mathcal D({\bf I}_{\boldsymbol\rho/c})$, and if a perturbation is detected then $\mathcal C(\mathcal R({\bf I}_{\boldsymbol\rho/c}))$ is evaluated for classification instead of $\mathcal C({\bf I}_{\boldsymbol\rho/c})$, where $\mathcal C(.)$ denotes the targeted network classifier.

\begin{figure*}[t] 
   \centering
   \includegraphics[width=3.7in]{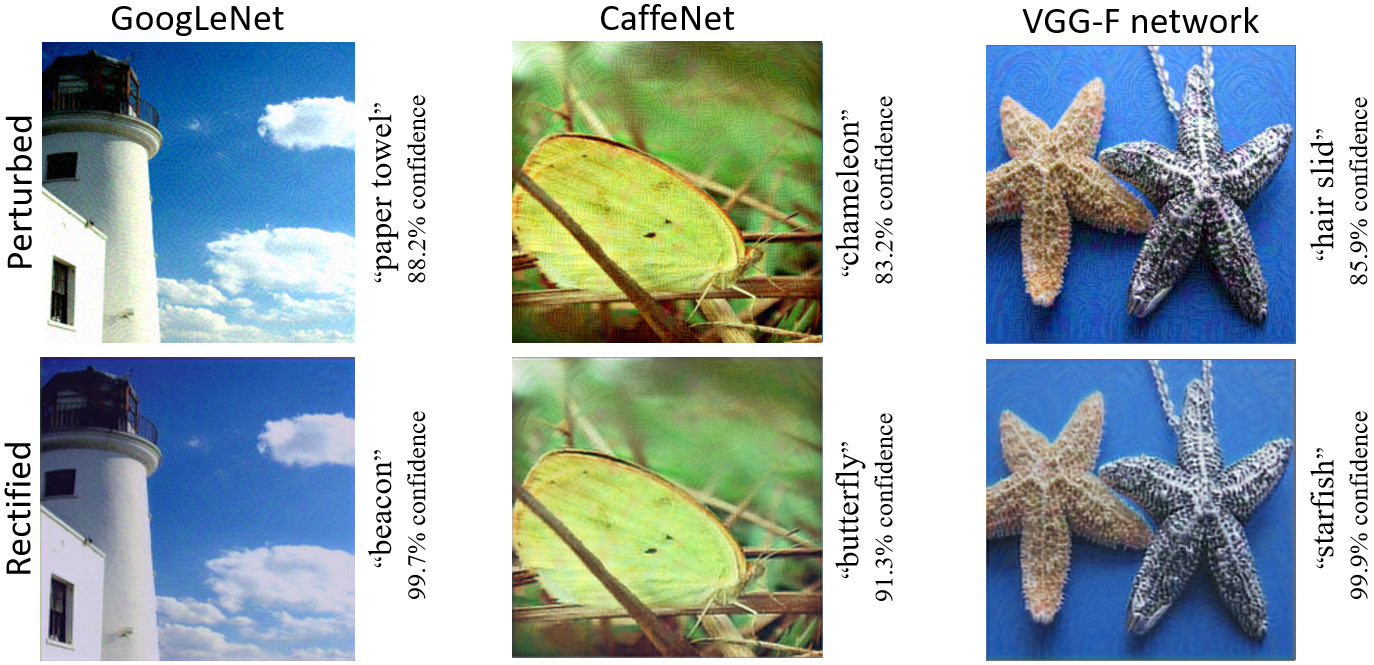} 
   \caption{Representative examples to visualize the perturbed images and their rectified version computed by the PRN. The labels predicted by the networks along the prediction confidence are also given. The examples are provided for the $\ell_{\infty}$-type perturbations. Please refer to the supplementary material of the paper for more examples.}  
   \label{fig:Expl}
   \vspace{-2mm}
\end{figure*}

\section{Experiments}
\label{sec:Exp}
We evaluated the performance of our technique by defending  CaffeNet~\cite{krizhevsky2012imagenet},  VGG-F network~\cite{chatfield2014return} and  GoogLeNet~\cite{szegedy2015going} against universal adversarial perturbations.
The choice of the networks is based on the computational feasibility of generating the perturbations for our experimental protocol. The same framework is applicable to other networks. Following Moosavi-Dezfooli~\cite{UniAdPert2017CVPR}, we used the ILSVRC 2012~\cite{krizhevsky2012imagenet} validation set of $50,000$ images to perform the  experiments.

\noindent{\bf Setup:} From the available images, we randomly selected $10,000$ samples to generate a total of $50$ image-agnostic perturbations for each network, such that $25$ of those perturbations were constrained to have $\ell_{\infty}$-norm equal to $10$, whereas the $\ell_2$-norm of the remaining $25$ was restricted to $2,000$.
The fooling ratio of all the perturbations was lower-bounded by 0.8. 
Moreover, the maximum dot product between any two perturbations  of the same type (i.e.~$\ell_2$ or $\ell_{\infty}$) was upper bounded by  0.15. 
This ensured that the constructed perturbations were  significantly different from each other, thereby removing any potential bias from our  evaluation.
From each set of the $25$ perturbations, we randomly selected $20$ perturbations to be used with the training data, and the remaining $5$ were used with the testing data.

We extended the sets of the training perturbations using the method discussed in Section~\ref{sec:synt}, such that there were total 250 perturbations in each extended set,  henceforth denoted as $\mathcal P^*_{\infty}$ and $\mathcal P^*_{2}$. 
To generate the training data, we first randomly selected $40,000$ samples from the available images and performed 5 corner crops of dimensions $224 \times 224 \times 3$ to generate $200,000$ samples.
For creating the adversarial examples with the $\ell_2$-type perturbations, we used the set $\mathcal P^*_{2}$ and randomly added perturbations to the images with 0.5 probability.
This resulted in $\sim 100,000$ samples each for the clean and the perturbed images, which were used to train the approach for the $\ell_2$-norm perturbations for a given network.
We repeated this procedure using the set  $\mathcal P^*_{\infty}$ to separately train it for the $\ell_{\infty}$-type perturbations. 
Note that, for a given targeted network we performed the training twice to evaluate the performance of our technique for both types of perturbations.

For a thorough evaluation, two protocols were followed to generate the testing data.
Both protocols used the unseen $10,000$ images that were perturbed with the $5$ unseen testing perturbations. 
Notice that the evaluation has been kept doubly-blind to emulate the real-world scenario for a deployed network.
For Protocol-A, we used the whole $10,000$ test images and randomly corrupted them with the $5$ test perturbations with a 0.5 probability. 
For the Protocol-B, we chose the subset of the $10,000$ test images that were correctly classified by the targeted network in their clean form, and corrupted that subset with 0.5 probability using the $5$ testing perturbations.
The existence of both clean and perturbed images with equal probability in our test sets especially ensures a fair evaluation of the detector.

\noindent{\bf Evaluation metric:} We used four different metrics for a comprehensive analysis of the performance of our technique. 
Let $\mathcal I_c$ and  $\mathcal I_{\boldsymbol\rho}$ denote the sets containing  clean and perturbed test images. Similarly, let $\widehat{\mathcal I}_{\boldsymbol\rho}$ and $\widehat{\mathcal I}_{\boldsymbol\rho/c}$ be the sets containing the test images rectified by PRN, such that all the images in $\widehat{\mathcal I}_{\boldsymbol\rho}$ were  perturbed (before passing through the PRN) whereas the images in $\widehat{\mathcal I}_{\boldsymbol\rho/c}$ were similarly perturbed with 0.5 probability, as per our protocol. 
Let $\overset{*}{\mathcal I}$ be the set comprising the test images such that each image is rectified by the PRN only if it were classified as perturbed by the detector $\mathcal D$. 
Furthermore, let $acc(.)$ be the function computing the prediction accuracy of the target network on a given set of  images. The formal definitions of the metrics that we used in our experiments are stated below:
\begin{enumerate} 
\item PRN-gain (\%) $= \frac{acc(\widehat{\mathcal I}_{\boldsymbol\rho}) - acc(\mathcal I_{\boldsymbol\rho})}{acc(\widehat{\mathcal I}_{\boldsymbol\rho})} \times 100$.
\item PRN-restoration (\%) $= \frac{acc(\widehat{\mathcal I}_{\boldsymbol\rho/c})}{acc(\mathcal I_c)} \times 100$.
\item Detection rate (\%) $=$ Accuracy of $\mathcal D$.
\item Defense rate (\%)  $= \frac{acc(\overset{*}{\mathcal I})}{acc(\mathcal I_c)} \times 100$.
\end{enumerate}

The names of the metric are in accordance with the semantic notions associated with them.
Notice that the PRN-restoration is defined over the rectification of both clean and perturbed images.
We do this to account for any loss in the classification accuracy of the targeted network incurred by the rectification of the clean images by the PRN.
It was observed in our experiments that unnecessary rectification of the clean images can sometimes lead to a minor (1 - 2\%) reduction in the classification accuracy of the  targeted network. Hence, we used a more strict definition of the restoration by PRN for a more transparent evaluation.   
This definition is also in-line with our underlying assumption of the practical scenarios where we do not know \emph{a prior} if the test image is clean or perturbed.

\setlength{\tabcolsep}{5.5pt}
\renewcommand{\arraystretch}{1}
\begin{table*}
\centering
\caption{Defense summary for the {\bf GoogLeNet}~\cite{szegedy2015going}:  The mentioned types of the perturbations (i.e.~$\ell_{2}$ or $\ell_{\infty}$)  are for the testing data.}
\begin{tabular}{lcc|ccc|ccc|ccc|c}  
\hline\hline
\multirow{4}{*}{{\bf Metric}} && \multicolumn{5}{c}{{\bf Same test/train perturbation type}} && \multicolumn{5}{c}{{\bf Different test/train perturbation type}} \\
\cline{3-7} \cline{9-13}
&& \multicolumn{2}{c}{$\ell_2$-type} && \multicolumn{2}{c}{$\ell_{\infty}$-type} && \multicolumn{2}{c}{$\ell_2$-type} && \multicolumn{2}{c}{$\ell_{\infty}$-type} \\
 \cline{3-4} \cline{6-7} \cline{9-10} \cline{12-13}
&& \multicolumn{1}{c}{Prot-A} & \multicolumn{1}{c}{Prot-B} && \multicolumn{1}{c}{Prot-A} & \multicolumn{1}{c}{Prot-B} && \multicolumn{1}{c}{Prot-A} & \multicolumn{1}{c}{Prot-B} && \multicolumn{1}{c}{Prot-A} & \multicolumn{1}{c}{Prot-B} \\
\cline{1-1}  \cline{3-4} \cline{6-7} \cline{9-10} \cline{12-13}
PRN-gain (\%)    		&& 77.0 & 77.1 && 73.9 & 74.2 && 76.4 & 77.0 && 72.6 & 73.4 \\
PRN-restoration (\%)   	&& 97.0 & 92.4 && 95.6 & 91.3 && 97.1 &92.7 && 93.8 & 89.3 \\
Detection rate (\%)    	&& 94.6 & 94.6 && 98.5 & 98.4 && 92.4 & 92.3 && 81.3 & 81.2 \\
Defense rate (\%)    	&& 97.4 & 94.8 && 96.4 & 93.7 && 97.5 & 94.9 && 94.3 & 91.6 \\
\hline
\end{tabular}
\label{tab:Google}
\vspace{-3mm}
\end{table*}
\setlength{\tabcolsep}{5.5pt}
\renewcommand{\arraystretch}{1}

\noindent{\bf Same/Cross-norm evaluation:} In Table~\ref{tab:Google}, we summarize the results of our experiments for defending  the GoogLeNet~\cite{szegedy2015going} against the perturbations.
The table summarizes two kinds of experiments. For the first kind, we used the same types of perturbations for testing and training. For instance, we used the $\ell_2$-type perturbations for learning the framework components (rectifier + detector) and then also used the $\ell_2$-type perturbations for testing. The results of these experiments are summarized in the left half of the table.
We performed the `same test/train perturbation type' experiments for both $\ell_2$ and $\ell_{\infty}$ perturbations, for both testing protocols (denoted as Prot-A and Prot-B in the table). 
In the second kind of experiments, we trained our framework on one type of perturbation and tested for the other. The right half of the table summarizes the results of those experiments. The mentioned perturbation types in the table are for the testing data.
The same conventions will be followed in the similar tables for the other two targeted networks below. Representative examples to visualize the perturbed and rectified images (by the PRN) are shown in Fig.~\ref{fig:Expl}. Please refer to the supplementary material for more illustrations.

From Table~\ref{tab:Google}, we can see that in general, our framework is able to defend the GoogLeNet very successfully against the universal adversarial perturbations that are specifically targeted at this network.
The Prot-A captures the performance of our framework when an attacker might have added a perturbation to an unseen image without knowing if the clean image would be correctly classified by the targeted network.
The Prot-B represents the case where the perturbation is added to fool the network on an image that it had previously  classified correctly. 
Note that the difference in the performance of our framework for Prot-A and Prot-B is related to the accuracy of the targeted network on clean images. For a network that is $100\%$ accurate on clean images, the results under Prot-A and Prot-B would match exactly.
The results would differ more for the less accurate classifiers, as also evident from the subsequent tables.

\setlength{\tabcolsep}{5.5pt}
\renewcommand{\arraystretch}{1}
\begin{table*}
\centering
\caption{ Defense summary for the {\bf CaffeNet~\cite{krizhevsky2012imagenet}: } The mentioned types of the perturbations (i.e.~$\ell_{2}$ or $\ell_{\infty}$)  are for the testing data.}
\begin{tabular}{lcc|ccc|ccc|ccc|c}  
\hline\hline
\multirow{4}{*}{{\bf Metric}} && \multicolumn{5}{c}{{\bf Same test/train perturbation type}} && \multicolumn{5}{c}{{\bf Different test/train perturbation type}} \\
\cline{3-7} \cline{9-13}
&& \multicolumn{2}{c}{$\ell_2$-type} && \multicolumn{2}{c}{$\ell_{\infty}$-type} && \multicolumn{2}{c}{$\ell_2$-type} && \multicolumn{2}{c}{$\ell_{\infty}$-type} \\
 \cline{3-4} \cline{6-7} \cline{9-10} \cline{12-13}
&& \multicolumn{1}{c}{Prot-A} & \multicolumn{1}{c}{Prot-B} && \multicolumn{1}{c}{Prot-A} & \multicolumn{1}{c}{Prot-B} && \multicolumn{1}{c}{Prot-A} & \multicolumn{1}{c}{Prot-B} && \multicolumn{1}{c}{Prot-A} & \multicolumn{1}{c}{Prot-B} \\
\cline{1-1}  \cline{3-4} \cline{6-7} \cline{9-10} \cline{12-13}
PRN-gain (\%)    			&& 67.2 & 69.0 && 78.4 & 79.1 && 65.3 & 66.8 && 77.3 & 77.7 \\
PRN-restoration (\%)   		&& 95.1 & 89.9 && 93.6 & 88.7 && 92.2 & 87.1 && 91.7 & 85.8 \\
Detection rate (\%)    		&& 98.1 & 98.0 && 97.8 & 97.9 && 84.2 & 84.0 && 97.9 & 98.0 \\
Defense rate (\%)    		&& 96.4 & 93.6 && 95.2 & 92.5 && 93.6 & 90.1 && 93.2 & 90.0 \\
\hline
\end{tabular}
\label{table:Caffe}
\end{table*}

In Table~\ref{table:Caffe}, we summarize the performance of our framework for the CaffeNet~\cite{krizhevsky2012imagenet}. Again, the results demonstrate a good defense against the perturbations.  
The final Defense-rate for the $\ell_2$-type perturbation for Prot-A is $96.4\%$.
Under the used metric definition and the experimental protocol, one interpretation of this value  is as follows.
With the defense wrapper provided by our framework, the performance of the CaffeNet is  expected to be $96.4\%$ of its original performance (in the perfect world of clean images), such that there is an equal chance of every query image to be perturbed or clean\footnote{We  emphasize that our evaluation protocols and metrics are carefully designed to analyze the performance in the real-world situations where it is not known apriori whether the query is perturbed or clean. }. 
Considering that the fooling rate of the network was at least $80\%$ on all the test perturbations used in our experiments, it is a good performance recovery.

\begin{table*}[h]
\centering
\caption{ Defense summary for the {\bf VGG-F} network~\cite{chatfield2014return}:  The mentioned types of the perturbations (i.e.~$\ell_{2}$ or $\ell_{\infty}$)  are for the testing data.}
\begin{tabular}{lcc|ccc|ccc|ccc|c}  
\hline\hline
\multirow{4}{*}{{\bf Metric}} && \multicolumn{5}{c}{{\bf Same test/train perturbation type}} && \multicolumn{5}{c}{{\bf Different test/train perturbation type}} \\
\cline{3-7} \cline{9-13}
&& \multicolumn{2}{c}{$\ell_2$-type} && \multicolumn{2}{c}{$\ell_{\infty}$-type} && \multicolumn{2}{c}{$\ell_2$-type} && \multicolumn{2}{c}{$\ell_{\infty}$-type} \\
 \cline{3-4} \cline{6-7} \cline{9-10} \cline{12-13}
&& \multicolumn{1}{c}{Prot-A} & \multicolumn{1}{c}{Prot-B} && \multicolumn{1}{c}{Prot-A} & \multicolumn{1}{c}{Prot-B} && \multicolumn{1}{c}{Prot-A} & \multicolumn{1}{c}{Prot-B} && \multicolumn{1}{c}{Prot-A} & \multicolumn{1}{c}{Prot-B} \\
\cline{1-1}  \cline{3-4} \cline{6-7} \cline{9-10} \cline{12-13}
PRN-gain (\%)    			&& 72.1 & 73.3 && 84.1 & 84.3 && 68.3 & 69.2 && 84.7 & 84.8 \\
PRN-restoration (\%)   		&& 93.2 & 86.2 && 90.3 & 83.2 && 88.8 & 81.2 && 91.1 & 83.3  \\
Detection rate (\%)    		&& 92.5 & 92.5 && 98.6 & 98.6 && 92.5 & 92.5 && 98.1 & 98.1 \\
Defense rate (\%)    		&& 95.5 & 91.4 && 92.2 & 87.9 && 90.0 & 85.9 && 93.7 & 89.1 \\
\hline
\end{tabular}
\label{tab:vggf}
\vspace{-3mm}
\end{table*}  

In Table~\ref{tab:vggf}, the defense summary for the VGG-F network~\cite{chatfield2014return} is reported, which again shows a decent performance of our framework.
Interestingly, for both CaffeNet and VGG-F, the existence of the $\ell_{\infty}$-type perturbations in the test images could be detected very accurately by our detector for the `different test/train perturbation type'. However, it was not the case for the GoogLeNet. 
We found that for the $\ell_{\infty}$-type perturbations (with $\xi =10$) the corresponding $\ell_2$-norm of the perturbations was generally much lower for the GoogLeNet ($\sim2,400$ on avg.) as compared to the CaffeNet and VGG-F ($\sim2,850$ on avg.). This made the detection of the $\ell_{\infty}$-type perturbations more challenging for the GoogLeNet.
The dissimilarity in these values indicate that there is a significant difference between the decision boundaries induced by the GoogLeNet and the other two networks, which is governed by the significant architectural differences of the networks. 

\noindent{\bf Cross-architecture generalisation:} With the above observation, it was anticipated that the cross-network defense performance of our framework would be better for the networks with the (relatively) similar architectures.
This prediction was verified by the results of our experiments in Tables~\ref{tab:CrossL2} and \ref{tab:Cross_LInf}. These tables show the performance for $\ell_{2}$ and $\ell_{\infty}$-type perturbations where we used the `same test/train perturbation type'.
The results are reported for protocol A. For the corresponding results under protocol B, we refer to the supplementary material. From these tables, we can conclude that our framework generalizes well across different networks, especially across the networks that have (relatively) similar architectures.
We conjecture that the cross-network generalization is inherited by our framework from the cross-model generalization of the universal adversarial perturbations.  Like our technique, any framework for the defense against these perturbations can be expected to exhibit similar characteristics.

\begin{table}
\centering
\caption{{\bf $\ell_{2}$-type cross-network defense} (Prot-A): Testing is done using the perturbations generated on the networks in the left-most column. The networks to generate the training perturbations are indicated in the second row. }
\begin{tabular}{ l|c|c|c }
  \hline\hline
  \multicolumn{4}{c}{PRN-restoration (\%)} \\
  \hline
   												& VGG-F &  CaffeNet & GoogLeNet\\
    \cline{2-4}
  VGG-F~\cite{chatfield2014return}   			& 93.2 	&  88.9 & 81.7 \\  
 CaffeNet~\cite{krizhevsky2012imagenet}   		& 91.3 	&  95.1	& 72.0 \\
  GoogLeNet~\cite{szegedy2015going}  			& 84.7 	&  85.9 & 97.0 \\

  \hline\hline
  \multicolumn{4}{c}{Defense rate (\%)} \\
  \hline
   												& VGG-F&  CaffeNet & GoogLeNet\\
    \cline{2-4}
  VGG-F~\cite{chatfield2014return}   			& 95.5	&  91.5 & 82.4 \\ 
  CaffeNet~\cite{krizhevsky2012imagenet}   		& 94.8 	& 96.2 	& 77.3 \\
  GoogLeNet~\cite{szegedy2015going}  			& 88.3 	& 87.3 	& 97.4 \\

  \hline
\end{tabular}
\label{tab:CrossL2}
\vspace{-3mm}
\end{table}

\begin{table}
\centering
\caption{{\bf$\ell_{\infty}$-type cross-network defense} summary (Prot-A). }
\begin{tabular}{ l|c|c|c }
  \hline\hline
  \multicolumn{4}{c}{PRN-restoration (\%)} \\
  \hline
   											& VGG-F &  CaffeNet & GoogLeNet\\
    \cline{2-4}
VGG-F~\cite{chatfield2014return}   			& 90.3 	&  86.9 & 74.1 \\ 
CaffeNet~\cite{krizhevsky2012imagenet}   	& 85.7 	& 93.6 	& 69.3 \\
GoogLeNet~\cite{szegedy2015going}  			& 85.9 	&  83.3 & 95.6 \\

  \hline\hline
  \multicolumn{4}{c}{Defense rate (\%)} \\
  \hline
   												& VGG-F &  CaffeNet & GoogLeNet\\
    \cline{2-4}
  VGG-F~\cite{chatfield2014return}   			& 92.2	& 88.9	& 74.8 \\  
  CaffeNet~\cite{krizhevsky2012imagenet}   		& 93.5 	& 95.2 	& 73.8 \\
  GoogLeNet~\cite{szegedy2015going}  			& 88.4 	&  85.4 & 96.4 \\
  \hline
\end{tabular}
\label{tab:Cross_LInf}
\vspace{-3mm}
\end{table}

\vspace{-3mm}
\section{Conclusion}
\label{sec:Conc}
\vspace{-2mm}
We presented the first dedicated framework for the defense against  \emph{universal adversarial perturbations}~\cite{UniAdPert2017CVPR} that  not only detects the presence of these  perturbations in the images but also rectifies the perturbed images so that the targeted classifier can reliably predict their labels.
The proposed framework  provides defense to a targeted model without the need of modifying it, which makes our technique highly desirable for the practical cases. 
Moreover, to prevent the potential counter-counter measures, it provides the flexibility of keeping its `rectifier' and `detector' components  secretive.
We implement the `rectifier' as a Perturbation Rectifying Network (PRN), whereas the `detector' is implemented as an SVM trained by exploiting the image transformations performed by the PRN.
For an effective training, we also proposed a method to efficiently compute image-agnostic perturbations synthetically.  
The efficacy of our framework is demonstrated by a successful defense of CaffeNet~\cite{krizhevsky2012imagenet}, VGG-F network~\cite{chatfield2014return} and GoogLeNet~\cite{szegedy2015going} against the universal adversarial perturbations.

\noindent{\bf Acknowledgement} This research was supported by ARC grant DP160101458. The Titan Xp used for this research was donated by NVIDIA Corporation.

\balance

{\small
\bibliographystyle{ieee}
\bibliography{egbib}
}

\end{document}